\documentclass{article}

\usepackage{arxiv}
\usepackage{natbib}
\usepackage[utf8]{inputenc} % allow utf-8 input
\usepackage[T1]{fontenc}    % use 8-bit T1 fonts
\usepackage{hyperref}       % hyperlinks
\usepackage{url}            % simple URL typesetting
\usepackage{booktabs}       % professional-quality tables
\usepackage{amsfonts}       % blackboard math symbols
\usepackage{nicefrac}       % compact symbols for 1/2, etc.
\usepackage{microtype}      % microtypography
\usepackage{lipsum}
\usepackage{graphicx}
\graphicspath{ {./images/} }

\title{Zero-Shot Pupil Segmentation with SAM 2: A Case Study of Over 14 Million Images}

\author{
 Virmarie Maquiling\footnote{\textsuperscript{*}These authors contributed equally to this work.} \\
  Human-Centered Technologies for Learning\\
  Technical University of Munich\\
  Munich Germany \\ 
  \texttt{virmarie.maquiling@tum.de} \\
   \And
 Sean Anthony Byrne\textsuperscript{*}\\
  Dipartimento di Elettronica, Informazione e Bioingegneria\\
  Politecnico di Milano\\
  Piazza Leonardo da Vinci \\
  Milano Italy\\
  \texttt{seananthony.byrne@polimi.it} \\
  \And
 Diederick C. Niehorster \\
  Lund University Humanities Lab \& Dept. of Psychology\\
  Lund University\\
  Lund, Sweden
  \texttt{diederick\_c.niehorster@humlab.lu.se} \\
  \And
 Marco Carminati \\
  Dipartimento di Elettronica, Informazione e Bioingegneria\\
  Politecnico di Milano\\
  Milano Italy\\
  \texttt{marco.carminati1@polimi.it} \\
  \And
 Enkelejda Kasneci \\
  Human-Centered Technologies for Learning\\
  Technical University of Munich\\
  Munich Germany \\ 
  \texttt{enkelejda.kasneci@tum.de} \\
}

\begin{document}
\maketitle
\begin{abstract}
We explore the transformative potential of SAM 2, a vision foundation model, in advancing gaze estimation and eye  tracking technologies. By significantly reducing annotation time, lowering technical barriers through its ease of deployment, and enhancing segmentation accuracy, SAM 2 addresses critical challenges faced by researchers and practitioners. Utilizing its zero-shot segmentation capabilities with minimal user input—a single click per video—we tested SAM 2 on over 14 million eye images from diverse datasets, including virtual reality setups and the worlds largest unified dataset recorded using wearable eye trackers. Remarkably in pupil segmentation tasks, SAM 2 matches the performance of domain specific models trained solely on eye images, achieving competitive mean Intersection over Union (mIoU) scores of up to 93\% without fine-tuning. Additionally, we provide our code and segmentation masks for these widely-used datasets to promote further research.
\end{abstract}

\begin{figure}
    \centering
    \includegraphics[width=\linewidth]{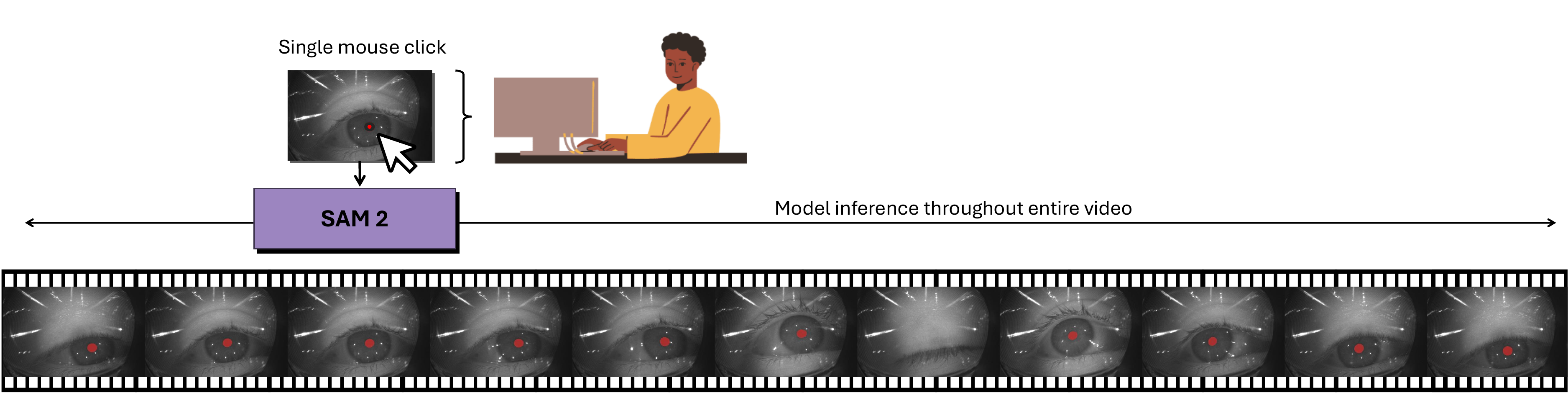}
    \caption{An illustration demonstrating the data annotation process with SAM 2: the user provides a single point prompt via a mouse click, and SAM 2 automatically handles the rest of the segmentation process. Optionally, the user can refine and add additional prompts in more difficult areas of the video to improve the model's output.}
    \label{fig:inference}
\end{figure}

\section{Introduction}
The increasing integration of eye tracking into technologies like virtual reality (VR) devices and smart glasses \cite{byrne2024lenses} has amplified the demand for robust gaze estimation systems, where a key task is the accurate localization of the pupil within an image or video \cite{kim2019nvgaze,maquiling2024zero}. Traditional methods for pupil localization—including thresholding and center of mass calculations \cite{nystrom2023amplitude, shortis1994comparison, perez2003precise} and ellipse-fitting algorithms \cite{santini2018pure, santini2018purest}—while effective in controlled environments, suffer catastrophic errors in the presence of noise such as occlusions or reflections, limiting their utility in real-world settings \cite{kothari2022ellseg, byrne2023leyeslightweightframeworkdeep}. To overcome these limitations, deep learning-based approaches have emerged as powerful alternatives, addressing issues plaguing traditional methods like blinks or reflections \cite{kim2019nvgaze} and improving the robustness and accuracy of pupil detection under challenging conditions \cite{fuhl2016pupilnet}. However, deploying these models requires vast amounts of annotated data and technical expertise. Data annotation can be very costly, requiring significant human labor depending on dataset size and complexity, and training gaze estimation models using supervised machine learning relies heavily on labeled data \cite{sun2017revisiting, sambasivan2021everyone}.

Foundation models represent a paradigm shift in artificial intelligence, transforming how people interact with, develop, and deploy deep learning models \cite{bommasani2021opportunities}. Characterized by vast numbers of trainable parameters and extensive training data, these models demonstrate impressive adaptability to downstream tasks and perform well on data distributions they have not encountered during training \cite{bommasani2021opportunities, kirillov2023segment}. They have lowered barriers to entry for integrating AI into workflows, simplifying the use of AI-powered tools and handling complex tasks that once required specialized models and custom datasets \cite{bommasani2021opportunities, zhou2023foundation}. Building on these advancements, \citeauthor{maquiling2024zero}~[\citeyear{maquiling2024zero}] showcased the potential of zero-shot vision foundation models in annotating eye tracking data by evaluating the Segment Anything Model (SAM) \cite{kirillov2023segment}, a vision foundation model released by Meta AI, on the OpenEDS datasets \cite{garbin2019openeds, palmero2020openeds2020}. However, SAM required at least one prompt per image, necessitating manual clicks on every image in the dataset—a time-consuming process. Its successor, SAM 2 \cite{ravi2024sam}, addresses this limitation by enabling a single prompt to propagate across an entire video, allowing the model to track and segment objects even with occlusions. This improvement drastically reduces the need for manual interaction, making the annotation process significantly more efficient.

This paper explores the potential of SAM 2 \cite{ravi2024sam}, in advancing gaze estimation research. Importantly, it addresses long-standing challenges in gaze estimation: the need for large annotated datasets, the labor-intensive process of feature annotation, the high barrier of expertise for developing custom models, and the difficulty of domain adaptation where models struggle to generalize across datasets \cite{kim2019nvgaze,kothari2022ellseg,byrne2023leyeslightweightframeworkdeep}. This feature is particularly relevant in gaze estimation, where models often fail due to variations in differences across participant physiology, recording setups, and environmental lighting conditions \cite{kim2019nvgaze, kothari2022ellseg, byrne2023leyeslightweightframeworkdeep}. 
To evaluate SAM 2, we deployed the model across diverse gaze estimation datasets, including VR environments and the world's largest unified public dataset of eye images captured with head-mounted devices \cite{fuhl2021teyed}. To evaluate and demonstrate SAM 2's ease of use, we limited the annotation to just one click per video, regardless of its length. A frame where the pupil was clearly visible was selected, and a single point prompt was placed near the center of the pupil. For the OpenEDS2019 dataset \cite{garbin2019openeds}, which consists of non-sequential eye images, we applied a single prompt to the entire dataset by taking the prompt from the first image in the training set and propagating it to the test and validation sets, covering a total of 152 different participants. We then assess SAM 2's segmentation performance using three key metrics: (1) the intersection-over-union (IoU) of the pupil masks to assess segmentation accuracy; (2) the ratio of frames where the pupil was not successfully tracked (when the pupil was visible) to the total number of frames (referred to as \textit{Pupil Lost}); and (3) the ratio of frames where a blink is correctly detected (i.e., the predicted mask is empty) to the total number of frames where the ground truth pupil mask is empty (referred to as \textit{Blink Detected}). Our annotation process employed SAM 2's smallest model, \texttt{SAM\_hiera\_tiny}, adapting the code released by SAM 2's authors \cite{ravi2024sam} so that it could handle arbitrarily long videos without preprocessing or running into memory issues. This improved version of SAM 2 used for this paper is available from \url{https://github.com/dcnieho/segment-anything-2} while the resulting masks can be downloaded from \url{https://zenodo.org/records/13911636}.

\section{Related Work}

\subsection{Foundation Models and the Segment Anything Models}

Models such as OpenAI's GPT series \cite{radford2018improving, radford2019language, brown2020language, achiam2023gpt} and Google's BERT \cite{kenton2019bert}, have transformed artificial intelligence by enabling versatile, zero-shot performance across a wide range of  downstream tasks \cite{bommasani2021opportunities}. With large parameter counts and trained on extensive data, these models can adapt effectively to new tasks without fine-tuning, making them broadly applicable beyond the field of Natural Language Processing (NLP) \cite{bommasani2021opportunities, zhou2023foundation}. This cross-domain adaptability has inspired similar advancements in other fields, with foundation models like TimeGPT \cite{garza2023timegpt} in time series forecasting, as well as DinoBloom \cite{koch2024dinobloom}, Nicheformer \cite{schaar2024nicheformer}, and others in medical research.

\begin{figure}
    \centering
    \includegraphics[width=\linewidth]{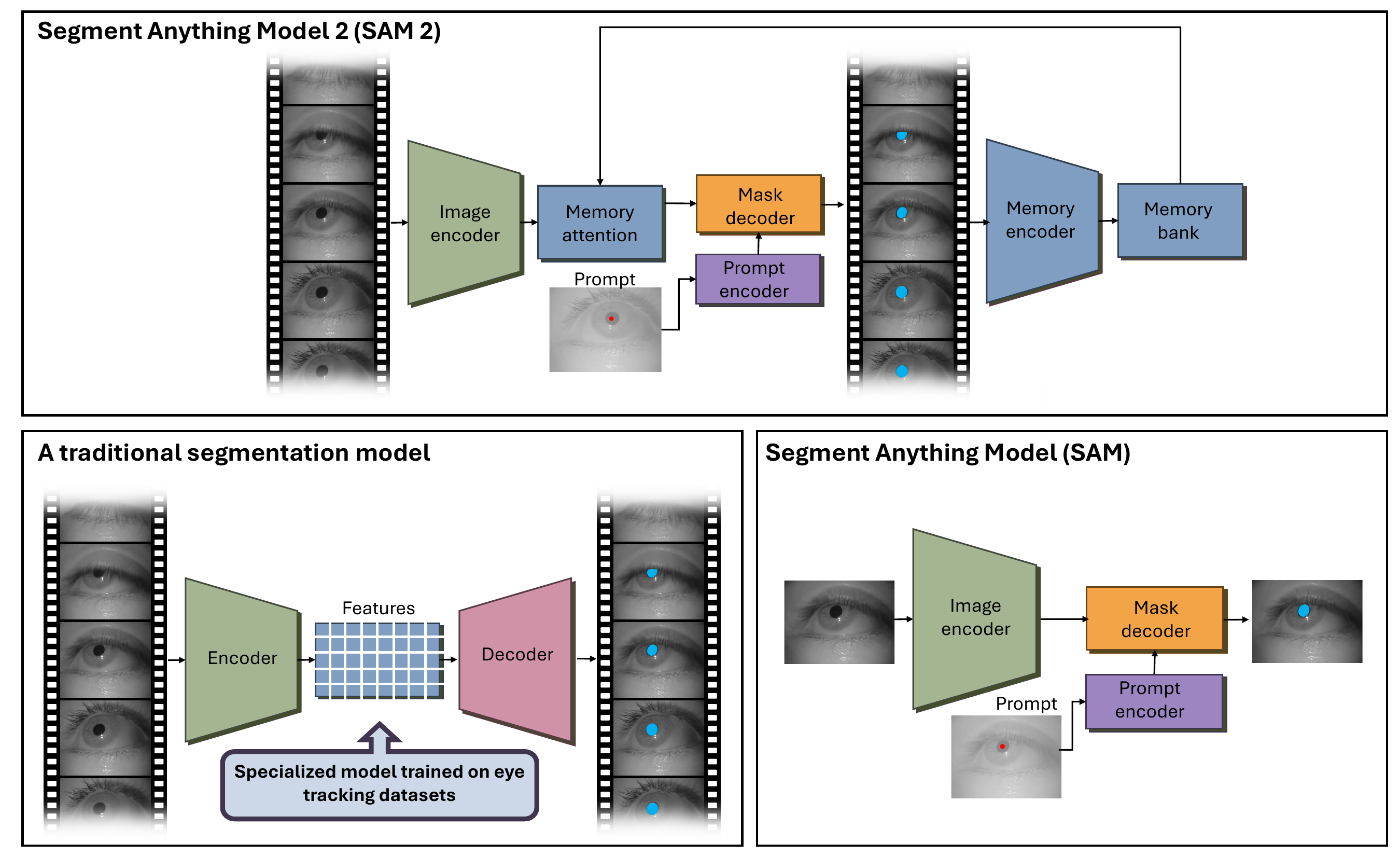}
    \caption{Comparison between Segment Anything Model 2 (top), a traditional segmentation model trained specifically on eye tracking datasets (left), and Segment Anything Model (right). The sample eye images are taken from the GW dataset \cite{kothari2020gaze}.}
    \label{fig:models}
\end{figure}

Foundation models have also achieved significant advances in computer vision, particularly in object recognition and segmentation. Notably, the Segment Anything Model (SAM) \cite{kirillov2023segment} represents a breakthrough in zero-shot image segmentation, enabling robust, general-purpose image segmentation across a wide range of visual inputs. SAM is trained on a massive diverse dataset (SA-1B), containing over one billion segmentation masks across 11 million images, making it capable of zero-shot segmentation across a wide range of image types. It operates by embedding both image and prompt inputs, such as points, bounding boxes, or text--and uses these embeddings to output segmentation masks. With its lightweight decoder and promptable structure, SAM can generate accurate masks with minimal input. Building on this, SAM 2 \cite{ravi2024sam} extends SAM's capabilities into video by introducing a memory-augmented architecture, enabling it to track and segment objects over time with higher accuracy and fewer user interventions. Unlike traditional frame-by-frame segmentation, SAM 2’s design allows users to provide a single prompt, such as a point or bounding box, to initiate segmentation across an entire video. Users can add additional prompts only where necessary, which the model incorporates as new input to further refine the segmentation in subsequent frames. Trained on the SA-V dataset--which includes over 35 million masks across 50,000 videos, SAM 2 excels at video object segmentation, reducing the need for prompts per frame and operating up to six times faster than SAM for static image tasks. This makes SAM 2 highly suitable for largescale applications. See Fig. \ref{fig:models} for a visualization of the architectural differences between SAM, SAM 2 and traditional specialist models. 

Since it was first introduced, the Segment Anything models has been applied in various fields such as medicine \cite{mazurowski2023segment,huang2024segment,zhang2024segment, ma2024segment}, remote sensing \cite{wang2024samrs, ding2024adapting, shankar2023semantic}, content creation \cite{yu2023inpaint, psychogyios2023samstyler} and autonomous driving \cite{yan2024segment, zhao2023enhancing}. Further, several extensions and improvements of the SAM model have been proposed, such as introducing speed-up alternatives \cite{zhao2023fast, zhang2023faster, zhang2023mobilesamv2}, extending it to 3D \cite{cen2023segment, shen2023anything, yang2023sam3d, guo2024sam2point}, object tracking in video \cite{yang2023track, cheng2023segment}, finetuning to specific domains \cite{ma2024segment} and so on. Since SAM 2 has been released only very recently, such adaptation have not yet appeared.

\subsection{SAM for Pupil Segmentation in Video-Based Eye Tracking}

In video-based eye tracking, segmentation of key eye regions such as the pupil, sclera, iris, and corneal reflections plays an important role in reliable gaze estimation. Central to many methods is the precise localization of the pupil center, which supports accurate gaze tracking \cite{kim2019nvgaze}. Current state-of-the-art methods fall into two categories: traditional approaches that rely on established image processing techniques, which use algorithmic rules to detect and segment eye features, and deep learning-based approaches. However, both have their limitations: traditional methods are susceptible to noise inherent in real-world scenarios, while deep-learning approaches, often require large, annotated datasets for training. 

To overcome these limitations, recent studies have explored the use of foundation models, which have the capacity to generalize across different datasets without the need for extensive retraining. Particularly, \citeauthor{maquiling2024zero}~[\citeyear{maquiling2024zero}] evaluated the efficacy of SAM on the OpenEDS datasets \cite{garbin2019openeds, palmero2020openeds2020} exploring different prompt strategies to segment three key eye features: the sclera, iris, and the pupil. While the iris and sclera showed less satisfactory results and required closer guidance using a combination of point prompts and a bounding box, they found that SAM could achieve competitive results for pupil segmentation with just the single point prompt indicating a random spot on the pupil, which compared well with specialist models trained on the same datasets. However, a prompt was needed for each image, making such a pipeline impractical when annotating large datasets such as the NVGaze \cite{kim2019nvgaze} datasets or the GIW dataset \cite{kothari2020gaze}, suggesting a need for additional steps in the pipeline or potentially fine-tuning SAM for eye-tracking datasets \cite{maquiling2024zero}. Later, \citeauthor{deng2024towards}~[\citeyear{deng2024towards}] explored a fully unsupervised approach to eye-region segmentation by using eye priors and image gradients to identify the pupil, iris, and sclera, with SAM assisting in refining boundaries. This method achieved near-supervised performance for pupil and iris segmentation, reaching approximately 90\% of benchmark accuracy.

\section{Methodology}

We evaluated SAM 2's performance on a diverse set of eye tracking datasets to test its generalizability across various domains. These datasets include both VR-based and mobile eye tracking environments, representing controlled and real-world settings. Specifically, we selected four VR-based datasets (including one synthetic) and three mobile eye tracking datasets captured in natural, uncontrolled environments. Two of these datasets are from the OpenEDS challenges \cite{garbin2019openeds, palmero2020openeds2020}, which focus on creating generalizable and robust semantic segmentations within VR settings. The synthetic NVGaze \cite{kim2019nvgaze} dataset includes its own pupil segmentations for gaze estimation. For the remaining datasets, ground truth segmentations were sourced from TEyeD \cite{fuhl2021teyed}, the world's largest unified public dataset of eye images taken with head-mounted devices.

Below is a brief description of the datasets used:

\begin{enumerate}
    \item \textbf{OpenEDS2019}~\cite{garbin2019openeds}: Contains 12,759 non-sequential images ($400 \times 640$ pixels) acquired from 152 participants using a VR head-mounted display (HMD) with eye-facing cameras at 200\,Hz under controlled lighting. Provides pixel-level annotations for the pupil, iris, and sclera.
    
    \item \textbf{OpenEDS2020}~\cite{palmero2020openeds2020}: Features eye-image sequences from 80 participants using a VR HMD at 100\,Hz. The Eye Segmentation Dataset includes 200 sequences sampled at 5\,Hz totalling to 29,500 images, of which 5\% are manually annotated ($640 \times 400$ pixels).
    
    \item \textbf{NVGaze}~\cite{kim2019nvgaze}: Comprises two datasets for near-eye gaze estimation under infrared illumination. The real-world dataset includes 264,279 images ($640 \times 480$ pixels) from 14 participants in a VR setting; the synthetic dataset contains 2 million images ($1280 \times 960$ pixels). We evaluated SAM 2 on both.
    
    \item \textbf{Labelled Pupils in the Wild (LPW)}~\cite{tonsen2016labelled}: Consists of videos from 22 participants recorded in everyday environments using a head-mounted eye tracker at 120\,Hz ($640 \times 480$ pixels), covering diverse lighting conditions and natural gaze distributions.
    
    \item \textbf{Gaze-in-Wild (GW)}~\cite{kothari2020gaze}: Provides naturalistic recordings from 19 participants performing everyday tasks with a mobile eye tracker at 120\,Hz ($640 \times 480$ pixels), including eye and head movements, infrared eye images, and scene imagery.
    
    \item \textbf{Dikablis datasets}: A combination of datasets from ElSe~\cite{fuhl2016else}, ExCuSe~\cite{fuhl2015excuse}, PNET~\cite{fuhl2016pupilnet}, and a driving study~\cite{kasneci2014driving}, compiled in TEyeD~\cite{fuhl2021teyed}. Recorded at 25\,Hz ($384 \times 288$ pixels), it features eye recordings from 30 participants.
\end{enumerate}

\section{Results}

\begin{figure}
    \centering
    \includegraphics[width=\textwidth]{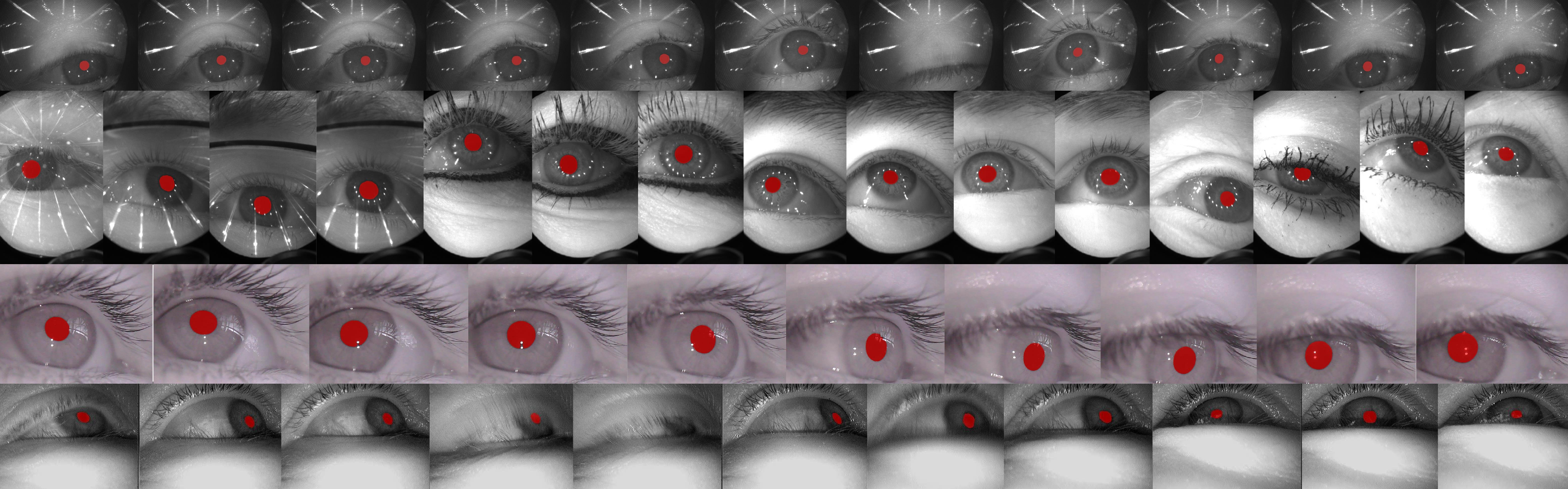}
     \caption{SAM 2 results on various VR-(first and second rows) and mobile eye tracking (third and last rows) datasets. Images are taken from the OpenEDS2019 \cite{garbin2019openeds}, OpenEDS2020 \cite{palmero2020openeds2020}, LPW \cite{tonsen2016labelled}, and the Dikablis datasets \cite{fuhl2021teyed}. SAM 2 handled occlusions remarkably well as observed in rows 1 and 4, and effectively segmented the pupil across a wide range of datasets, showing its robustness to different eye tracking conditions.}
    \label{fig:gallery}
\end{figure}

\begin{table}[htbp]
    \centering
    \begin{tabular}{l|l|l|l|ccc}
        \hline
         & \textbf{OpenEDS} &  \textbf{EllSeg-Gen}&\textbf{TEyeD} & \multicolumn{3}{c}{\textbf{SAM 2}} \\
         % \cline{4-6}
         & \textbf{Mean IoU} & \textbf{ Mean IoU}&\textbf{Mean IoU} & \textbf{Mean IoU} & \textbf{Pupil Lost} & \textbf{Blink Detected} \\
        \hline
        \textbf{OpenEDS2019} & 0.9528 &  0.956&-- & 0.8997 & 0.0075 & 0.9412 \\
        \textbf{OpenEDS2020} & 0.9517 &  --&-- & 0.9233& 0.0104& 0.9375\\
 \textbf{NVGaze (Synthetic)}      & -- &  0.982&--& 0.9259 & 0.0028 &0.9602 \\
 \textbf{NVGaze (Real)}      & -- &  --&0.65   & 0.9079 & 0.0876 &0.9889 \\
        \textbf{LPW}         & -- &  --&* & 0.9023 & 0.0940 & 0.7127 \\
        \textbf{GW}          & -- &  --&*   &  0.9212    & 0.0184     & 0.8513     \\
        \textbf{Dikablis}    & -- &  --&*   & 0.8835& 0.1846& 0.9483\\
        \hline
    \end{tabular}
    \caption{Performance of SAM 2 on multiple datasets compared to leaderboard scores published in OpenEDS2019 and OpenEDS2020 challenge pages, and baseline result from TEyeD using leave-one-out cross validation for each eye tracker. The asterisk (*) indicates datasets where a single baseline value was reported for all four datasets. As TEyeD provided the ground truth segmentation for NVGaze (real), LPW, GW, and the Dikablis datasets, no baseline mean IoU could be extracted from the original papers. Similarly, no baseline IoU was reported for the NVGaze synthetic dataset, therefore, we extracted the mIoU score from \cite{kothari2022ellseg} who evaluated their model on NVgaze. Their mIoU score for OpenEDS2019 has been included as well.}
    \label{tab:results}
\end{table}

Table\ref{tab:results} summarizes SAM 2's performance metrics—mean IoU, pupil lost rate, and blink detection rate—compared to the leaderboard scores\footnote{https://eval.ai/web/challenges/challenge-page/353/leaderboard/1002}\textsuperscript{,}\footnote{https://eval.ai/web/challenges/challenge-page/603/leaderboard/1680} of OpenEDS and the baseline score of TEyeD. For instance, on the OpenEDS2019 dataset, SAM 2 achieved a mean IoU of 89.97\%, slightly lower than specialist models trained on OpenEDS, which reached at most 95.6\%, with a pupil lost rate of 0.75\% and a blink detection rate of 94.12\% \cite{garbin2019openeds}. Similarly, for OpenEDS2020, SAM 2 attained a mean IoU of 92.33\%, compared to the specialist model's 95.17\% \cite{palmero2020openeds2020}.

On the NVGaze datasets \cite{kim2019nvgaze}, SAM 2 performed well on synthetic data with a mean IoU of 92.59\% and on real data with 90.79\%, despite a higher pupil lost rate of 8.76\% on the real data. In mobile datasets, SAM 2 achieved mean IoUs of 90.23\% on LPW \cite{tonsen2016labelled} and 92.12\% on GW \cite{kothari2020gaze}, with higher pupil lost rates due to increased noise and visual obstructions, as evidenced by the Dikablis datasets' pupil lost rate of 18.46\% \cite{fuhl2016else, fuhl2015excuse, fuhl2016pupilnet, kasneci2014driving, fuhl2021teyed}.

Overall, SAM 2 \cite{ravi2024sam} performed well on both VR and mobile eye tracking datasets, with VR datasets showing higher performance likely due to more controlled environments. Although SAM 2 slightly underperformed on the nonsequential OpenEDS2019 dataset—composed of individual images rather than continuous video frames—it still demonstrated the ability to generalize across multiple datasets. Notably, SAM 2 outperformed TEyeD's top-performing model, which achieved a mean IoU of just 65\% \cite{fuhl2021teyed}, and delivered competitive results compared to \cite{kothari2022ellseg}, where a mean IoU of 98.2\% was reported on NVGaze's synthetic dataset. It is important to highlight that their model was trained on several datasets, including NVGaze's training set, before being tested on the NVGaze test set. In contrast, SAM 2 achieved these results without any fine-tuning or sacrificing a subset of the datasets for training.

A major advantage of SAM 2 is its minimal human guidance requirement—just a single click per video to indicate that it should segment the pupil—compared to traditional manual annotation that would require thousands of clicks. For example, SAM 2 required only one click for the entire OpenEDS2019 dataset of 12,759 images. Similar reductions were observed across other datasets: 200 clicks for the 29,500 images of the OpenEDS2020 dataset, 66 clicks for LPW's 130,856 images, 54 clicks for NVGaze's 2,264,279 images, 423 clicks for the Dikablis datasets' more than 5.6 million images, and 148 clicks for GW's 6 million images. This significant reduction in human labor highlights SAM 2's efficiency in annotating large datasets with minimal input while achieving high performance in pupil tracking without model fine-tuning or specialized training. Moreover, SAM 2's inference process did not require powerful GPUs, making it feasible for use with low-end hardware. This level of accessibility was not possible before SAM 2 and demonstrates how vision foundation models can democratize large-scale data annotation in eye tracking research.

\section{Discussion}
The quality of the dataset significantly impacts SAM 2's performance. For instance, datasets with clear eye images, minimal noise and high resolution, such as OpenEDS and NVGaze, yielded higher IoU scores and lower pupil loss rates. However, SAM 2 encountered more difficulty in noisier datasets, like the Dikablis datasets, resulting in more pupil tracking failures. 

In terms of human interaction, the effort required is mostly limited to monitoring the quality of the predicted masks and adding additional prompts only when necessary. While we limited our evaluation to a single point prompt per video, SAM 2 supports various other prompt types, such as multiple positive or negative point prompts (where the signs indicate areas that SAM 2 should and should not include in its segmentation) and bounding box prompts, offering flexibility for more complex and noisier datasets that require additional guidance. Below we highlight several practical lessons from conducting our study for researchers interested in implementing SAM 2 for eye tracking data segmentation tasks:

\subsection{Practical Lessons Learned} 
\begin{enumerate}
    \item \textbf{Significant Time Savings:} SAM 2 significantly reduced the time necessary to annotate entire datasets. The datasets in this study were annotated by two of the authors within a couple of days, demonstrating SAM 2's efficiency in annotating large volumes of data quickly. Importantly, these authors spent most of the time waiting for the model to finish producing its segmentation for the datasets, and only very little time setting up the model and checking its output.

    \item \textbf{Reduced Technical Barrier to Entry:} SAM 2 not only saves time in obtaining segmentation masks but also greatly simplifies the process for non-experts. Instead of developing custom pupil segmentation models, users can run SAM 2 with just a few lines of code. This lowers the barrier to entry, enabling more people to develop gaze estimation pipelines.
    
    \item \textbf{Minimal Human Interaction:} The annotation process required minimal human involvement beyond preparing the prompts and finding appropriate frames where the pupil is visible. SAM 2 handled the actual annotation, while the user only needs to perform quality checks and refine prompts when necessary. 
    
    \item \textbf{Dealing with Noisy Data} While SAM 2 performed well in most cases, a decrease in performance was observed in noisier datasets. To mitigate this, further refining of prompts (e.g. using a more appropriate prompt strategy or adding prompts on more difficult frames) is necessary--although this still involves less  human effort compared to traditional annotation processes.
    
    \item \textbf{Low Hardware Requirements} SAM 2's low GPU requirements make it accessible for researchers with limited computational resources. We used both a high-end NVIDIA A100 (80GB VRAM) and a GeForce RTX 4090 (24 GB VRAM), achieving compute times of up to 40 frames per second (fps) for both. Impressively, SAM 2 also ran on more budget-friendly GPU's such as the RTX 4060 Ti (16 GB VRAM) delivering around 12 fps, and was even functional on laptop-class GPU's, albeit at significantly lower frame rates of just a few fps. However, it is worth noting that inference performance of the model appeared to be limited by CPU and not GPU performance in most of these cases. 
    
    \item \textbf{Strong Generalization} SAM 2 demonstrated robust performance across a diverse set of eye tracking datasets including VR, mobile, and even synthetic data, despite not being specifically trained on eye tracking data. 
    
\end{enumerate}

\subsection{Open Challenges, Limitations \& Future Work}

While SAM 2 excelled in pupil segmentation, challenges remain with segmenting less distinct eye features like the iris and sclera. To explore this, we evaluated SAM 2's performance on the OpenEDS2020 dataset, where it achieved an mIoU of 76.53\% for the iris and only 7.36\% for the sclera with a single box prompt. Fine-tuning SAM 2 on specific eye features, especially under varying conditions like lighting, reflections, and noise, could improve performance by reducing its reliance on ideal conditions and simple prompts.

Additionally, alternative prompting strategies, such as using bounding boxes or multiple point prompts, may yield better results in challenging cases. A limitation of our study is that we did not explore different prompt strategies, opting for a single point prompt to highlight the simplicity of annotation with SAM 2. While a single prompt proved to be effective for pupil segmentation in many cases, other prompts may improve results in difficult cases or for less well defined features.

Another consideration is the possibility that SAM 2 may have encountered similar images during training as all the datasets we have used are open to the public. Future work should test SAM 2 on completely novel datasets to validate its generalization capabilities. Additionally, as eye tracking moves to consumer devices, a key challenge will be adapting SAM 2 for low-power hardware like smart glasses \cite{zhang2023faster}, making it crucial to balance performance with reduced computational requirements for real-time applications in VR, AR devices.

\subsection{Privacy and Ethics}
The ease of use provided by SAM 2 significantly reduces the need for manual annotation in large datasets, which is generally a positive development for eye-tracking research. However, this increased efficiency also has the potential to scale up the collection and processing of personal data, particularly sensitive biometric information. As self-annotated eye-tracking datasets become more feasible with SAM 2, it is crucial that data sovereignty and informed consent protocols are rigorously upheld to ensure ethical compliance. Researchers must prioritize transparency and safeguard user data, especially when dealing with large-scale, potentially sensitive datasets.

\section{Conclusion}

In this study, we assessed the practical segmentation capabilities of the SAM~2 Vision Foundation model. Using SAM~2, we efficiently annotated over 14 million pupil images across multiple datasets with just a few click prompts per dataset, significantly streamlining traditional annotation workflows. Our findings show that foundation models like SAM 2 effectively address key challenges in eye tracking research: data annotation, domain adaptation, and reducing training data requirements. Notably, SAM 2 achieves robust performance without fine-tuning, offering a user-friendly and accurate solution compared to its predecessor, SAM. Its ability to annotate entire datasets with minimal human input makes it suitable practically for large-scale applications. Further, it could be used to standardize gaze estimation across datasets, ensuring fair comparisons. This standardization not only supports reproducibility but also facilitates cross-study evaluations in gaze estimation research. 

This work highlights the potential for general-purpose models to benefit other HCI fields where extensive labeled data is needed. As these models advance, we anticipate continued progress in both eye tracking research and broader human-computer interaction applications.
\bibliographystyle{ACM-Reference-Format}
\bibliography{references}

\end{document}